\documentclass[letterpaper, 10 pt, conference]{ieeeconf}  

\IEEEoverridecommandlockouts                              

\usepackage{booktabs}
\usepackage{multirow}
\usepackage{amsmath}
\usepackage{graphicx}
\usepackage{algorithm}
\usepackage{}
\usepackage{algorithmic}
\usepackage[colorlinks=true, allcolors=blue]{hyperref}

\title{\LARGE \bf
Improving Policy Learning from Delayed Interventions
}

\author{Xiru Zhu, David Meger, Gregory Dudek
\thanks{McGill University, Computer Science}
}

\begin{document}

\maketitle

\begin{abstract}
In this work, we explore the idea of expert intervention delays in human-supervised robotic systems for imitation learning system. Previous works in imitation learning have primarily examined the impact of suboptimal experts, where the expert policy makes mistakes, but assumed expert interventions occur immediately after errors are made. Although delayed expert intervention can be viewed as a form of noise that might improve policy robustness, our results show that learning from delayed expert interventions is overall less effective than learning from non-delayed expert interventions; delayed expert interventions are less effective at demonstrating error avoidance behaviors.

To address this, we propose Learning from Delayed Intervention (LfDI), which learns a cost function based on where and when the expert intervenes, combined with limited information about the extent of intervention delays provided by the expert. This cost function is then used to infer additional supervision for states without expert data, guiding the policy to avoid states that would likely trigger interventions. We evaluate our method, Learning from Delayed Intervention (LfDI), on simulated environments and show that it outperforms other methods that do not account for intervention delays. 
\end{abstract}

\section{INTRODUCTION}

Human-gated imitation learning iteratively improves a policy by sampling trajectories using the current policy, where the expert takes control when the policy makes mistakes\cite{kelly2019hg}. Several studies have examined how to improve the performance of imitation learning methods when experts provide suboptimal actions~\cite{brown2020better, xu2022discriminator, yu2023offline}. In this work, we instead examine how delays in expert interventions affect learning for human-gated imitation learning. We assume that experts do not always react immediately to errors: they may be distracted, or some small errors may only become apparent after accumulating over time, leading to delayed interventions. Furthermore, as the policy improves, experts may increasingly trust the policy to recover from minor mistakes~\cite{xu2015optimo}, which can introduce additional delays if the recovery fails. Thus, delays in expert intervention are likely common, yet they remain largely unaddressed in research.

The challenge with delayed intervention is that expert demonstrations may not teach the policy how to correct errors early, but instead show recovery only after significant errors accumulate. For example, consider an autonomous robot veering off its intended path. If the expert intervenes just before the robot crashes into a tree, the agent learns how to avoid crashing into trees. However, it may not learn how to avoid veering off course in the first place. Consequently, additional interventions, which can be costly are required until more timely interventions demonstrate how to prevent the initial error. As we will show later, training with delayed expert interventions can significantly degrade policy performance.

\begin{figure}
\includegraphics[scale=0.5]{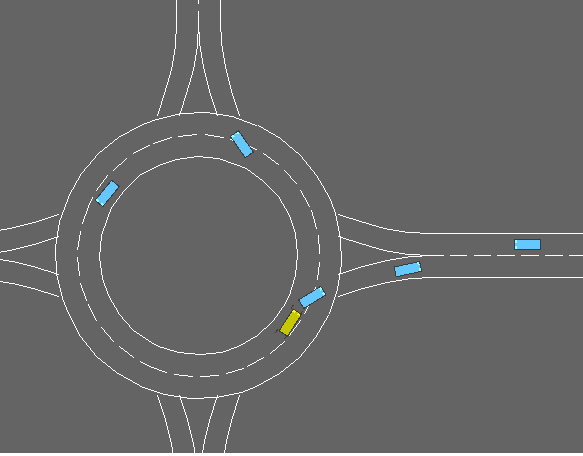}
    \caption{To evaluate our proposed solution, we experimented on several environments in the highway-env\cite{highway-env}. The \textbf{Roundabout} environment is configured such that the vehicle starts on the south route outside the roundabout and must enter the roundabout and exit to the north route to complete the task. In this figure above, the expert intervened and braked, preventing the controlled vehicle(yellow) from crashing into another vehicle.}
    \label{fig:roundabout}
\end{figure}

To learn from delayed interventions, we extend the work by Spencer et al \cite{spencer2020learning, spencer2022expert}. In their work, they proposed learning a binary classification cost function from states without interventions and states immediately preceding an intervention within a small time window \cite{spencer2020learning}. They assume that states preceding an intervention are bad states to be avoided. The cost function can then be used to improve policy learning. Delayed expert interventions, however, require considering a much larger time window of potentially  bad state-action pairs. With longer intervention delays, more states are visited before the expert takes control, making it difficult to determine which state-action pairs caused the intervention. As a result, some state-action pairs within the window may be safe, while other unsafe pairs may occur in trajectories that never trigger an intervention. Thus, our goal is to infer which actions within the time window led to the intervention, without over-penalizing actions that did not. Given the wide range of possible intervention delays, we assume access to some information about the level of delay for each intervention. While experts may introduce intervention delays for various reasons, they are generally aware of the extent of their delays in hindsight. We do not assume access to very precise delay estimates, as humans are often inaccurate at providing direct numerical values \cite{mandel2017add}. 

In this paper, we address delayed expert interventions in human gated imitation learning. Our contributions are:

\begin{itemize}
\item We formalize the problem of delayed expert interventions in human-gated imitation learning and analyze the challenges they introduce.
\item We introduce \textit{Learning from Delayed Interventions} (LfDI), which leverages the timing and locations of expert interventions to improve learning from delayed interventions.
\item We evaluate LfDI on \textit{highway-env} (Highway, Roundabout, Intersection)\cite{highway-env} and MuJoCo Humanoid\cite{todorov2012mujoco}, showing reduced crashes without sacrificing performance.
\end{itemize}

\section{Related Work}
\label{sec:related_work}
\subsection{Active Imitation Learning} Imitation learning (IL), the process of learning from expert demonstrations, which can be implemented using behavior cloning(BC), is a powerful approach to learning a policy from expert demonstrations. With BC, the agent directly learns the state-to-action mapping from expert demonstrations. For more complex problems, training BC typically requires a large number of demonstrations to generalize effectively. Given an expert policy $\pi_E$ and trajectories $\tau_{E}$ obtained from executing $\pi_E$, BC seeks to minimize the difference between learned policy and the expert policy, expressed as $J(\pi, \pi_E|\tau) = \sum_{t=0}^T l(\pi(s_t), \pi_E(s_t))$ \cite{laskey2017dart}.  

BC, however, can suffer from catastrophic failures due to compounded errors as the agent deviates from the training state distribution. DAgger addresses this problem by incorporating human demonstrations of how to recover from mistakes \cite{ross2011reduction}. In DAgger, the expert provides actions at randomly sampled states, but the agent does not execute these actions, allowing it to explore and learn recovery strategies \cite{ross2011reduction}. A limitation of DAgger is that it does not give the human expert sufficient control \cite{kelly2019hg}. Expert actions are only executed with some probability determined by $\beta$, which reduces feedback to the expert and may affect the quality of intervention data. To address this, Kelly et al. \cite{kelly2019hg} proposed an intervention-based version of DAgger in which the expert can take full control of the agent at any time.
\begin{figure}
\includegraphics[scale=0.32]{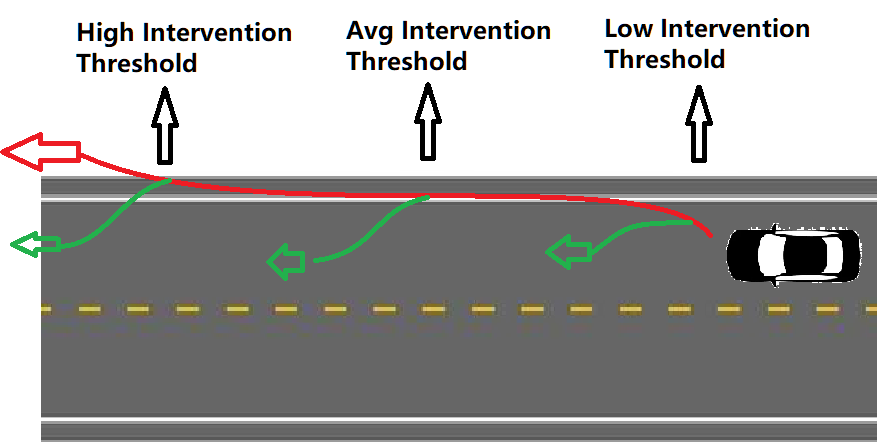}
    \caption{Consider a vehicle attempting to stay on the road. During data sampling, the current policy may make mistakes and gradually drift from the intended path. Depending on the intervention threshold $\alpha$, the expert may intervene at different points. When $\alpha$ is low, the acceptable total error over the window $w$ is small, and the expert intervenes promptly after errors occur. In contrast, when the expert is distracted, their $\alpha$ is higher, and interventions are delayed.}
    \label{fig_supervision_level}
\end{figure}

Building upon HG-dagger, several intervention-based imitation learning methods have been proposed to extract more information from human interventions, including cases where no intervention occurs \cite{spencer2020learning, spencer2022expert, senft2019teaching, chisari2022correct}. Spencer et al. proposed learning from both the timing of interventions and the absence of interventions. They suggest learning a cost function that penalizes states immediately preceding interventions while rewarding states where no intervention occurs. The BC policy is then trained to minimize both this intervention cost function and the standard behavior cloning loss. Their approach, however, does not address cases where expert interventions are delayed. This creates a much larger window of potentially simply ``bad" states, which motivates our proposed extension.

\subsection{Human Preference Learning for Robotics}
In many problems, defining a reward function can be challenging due to its complexity. Instead of manually specifying a reward function, demonstrations can be used to either apply inverse reinforcement learning to infer the reward function or employ imitation learning to replicate the demonstrated behaviors. Obtaining demonstrations can, however, be difficult, particularly for robots that are difficult for humans to operate or for tasks where skilled human experts are scarce. Directly providing human feedback as a reward can be costly. Moreover, humans generally excel at offering relative judgments rather than absolute ones \cite{daniel2014active, macglashan2017interactive, christiano2017deep}. An alternative approach is to present pairs of trajectories to human experts to learn their relative preferences, which forms the basis of reinforcement learning from human preferences.

Human Preference Reinforcement Learning(HPRL) was first proposed by \cite{furnkranz2012preference} and 
\cite{daniel2014active} for robotics problems. \cite{christiano2017deep} conducted a large-scale study on HPRL using Eq.~8 for Atari games, based on binary human preferences. This did require a large amount of human labeling: the reward function had to be trained using preferences from sampled trajectory pairs of the current policy, necessitating frequent human labeling, with over 5,500 labels per policy required to train ATARI games. 

Given the limited data available, HPRL research tends to focus on reducing the number of labels required, such as by using data augmentation \cite{park2022surf} or improved exploration \cite{liang2022reward}. In a similar vein, Hejna et al\cite{hejna2023few} propose using data from multiple tasks to learn a meta-preference reward function that can be adapted to the current task with only a few preference labels. Other research focuses on improving reward function learning, such as by adding a state prediction loss to the reward function \cite{liu2023efficient}. 

We note that most human preference RL research in robotics uses an oracle reward function instead of actual human preferences \cite{lee2021pebble, sadigh2017active, park2022surf, cao2021weak} due to cost constraints, except for some qualitative evaluations, such as teaching a robot to lift its left leg. Oracle functions are deterministic, and this lack of noise can be detrimental to learning true preferences, as argued by \cite{laidlaw2021uncertain, barnett2023active}. Instead, Laidlaw et al suggest that incorporating uncertainty can improve preference learning, especially in regions near decision boundaries, and can enhance learning of softer reward functions. Xue et al \cite{XueAnYanXu24} proposed improving human preference reward learning from actual human preferences by stabilizing the reward function with regularization and constraints. Finally, human preference RL has been applied to simulations and real robots, such as for social navigation \cite{mehta2024unified}, or combined with demonstrations and human intervention \cite{wang2022feedback}.
\subsection{Delays in Robotics}
One form of delay commonly studied in robotics is latency and delays in teleoperated systems \cite{farajiparvar2020brief}. In these settings, the robot is primarily controlled by an expert, with less focus on supervising an autonomous policy. Predictive methods are sometimes employed to approximate expert actions when delays occur. This issue is particularly relevant for teleoperated robots in remote environments, such as space, where consistent intervention delays are unavoidable \cite{seo2024exploring}. Delays are also examined in human–robot interaction, where Pascher et al.~\cite{pascher2023communicate} explore strategies for managing interactions when the robot experiences delays and for improving communication when the robot cannot respond immediately. However, these methods deal with delays that are more predictable, which is less the case for intervention delays. 

\begin{figure}
\includegraphics[scale=0.28]{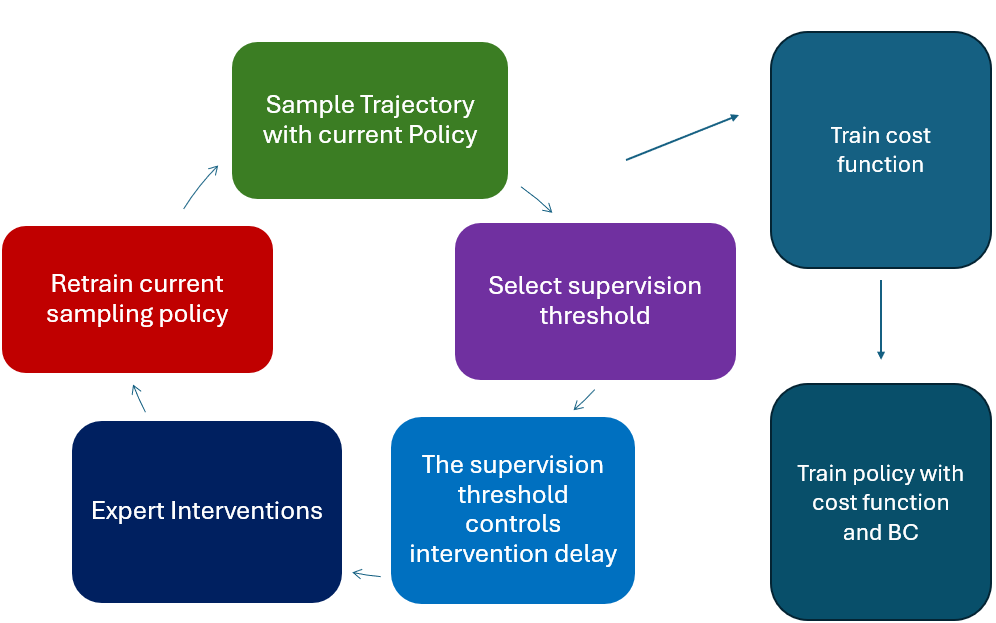}
\caption{Learning from Delayed Interventions (LfDI) has two phases: data sampling and training. During sampling, expert interventions occur when the cost exceeds a threshold $\sigma$, with higher thresholds allowing longer delays and tolerating more errors. After collecting trajectories, the policy is retrained with the new data. Finally, the cost function is trained, and the policy is retrained from scratch with Equation~\ref{total_loss_idea_1}.}
\label{delayed_intervention_data_sampling}. 
\end{figure}
\section{Learning from Delayed Intervention}
\label{sec:model}
This section presents our human-gated imitation learning framework for handling delayed expert interventions. We first formalize intervention delays and define the mechanism that triggers expert interventions. We then describe how trajectories are collected and how they are incorporated into the training dataset. Finally, we present the learning objective and training procedure used to optimize the agent’s policy with delayed intervention data. 

Our human-gated imitation learning framework is based on HG-DAgger from Kelly et al.~\cite{kelly2019hg}, where in their work, interventions are human-gated. In this work, the expert intervenes when the sum of errors within a time window $w$ exceeds an intervention threshold, similar to the system in Spencer et al.~\cite{spencer2020learning}. Thus, an intervention is triggered based on accumulated error tolerance, which we refer to as the intervention threshold $\alpha$. The higher the intervention threshold, the more delayed an intervention will be because more error must accumulate to reach the threshold. However, interventions can also be triggered immediately if a critical error occurs (i.e., error exceeds threshold immediately). The intervention threshold formalizes the human gating for interventions in our framework. When the expert takes control, they can execute as many steps as necessary before returning control to the agent when it is safe. A focused expert has a low $\alpha$, reacting promptly with minimal delay, whereas a distracted expert has a higher $\alpha$, responding only to critical errors or failing to react entirely, as shown in Figure~(\ref{fig_supervision_level}).

To compute when the agent's actions exceed $\alpha$, we assume that the expert has an internal optimal cost function, $c^*(s, a)$, which outputs a scalar. We also assume that the expert considers only a window of $w$ past states when calculating the internal cost. For example, with $w=30$, an intervention at $t=35$ is triggered when accumulated errors from $t=5$ to $t=35$ exceed $\alpha$.  An intervention, which we denote as $b$ for simplicity, is triggered as follows, where $\sigma_{t-w:t}$ denotes the trajectory segment from $t-w$ to $t$:
\begin{equation}
    c^*(\sigma_{t-w:t}) = \sum_{i=t-w}^{t}c^*(s_i, a_i)
\end{equation}
\begin{equation}
     b(\sigma_{t-w:t}, \alpha) =
    \begin{cases}
            1, &         \text{if } c^*(\sigma_{t-w:t}) > \alpha,\\
            0, &         \text{otherwise }
    \end{cases}
\end{equation}

This assumption is based on the work of Spencer et al.\cite{spencer2020learning}, which assumes that the states leading to an intervention are to be avoided. This is supported by the idea that, as the policy makes mistakes, it deviates from the state distribution of the training data, leading to compounding errors \cite{ross2011reduction}. For this work, we do not account for errors extending beyond the time window $w$.

Let $T$ be the set of trajectories with delayed interventions collected by the sampling policy $\pi$. Each trajectory $\tau \in T$ is defined as a list of tuples, $\tau = [(s_0, a_0, b_0, \alpha_0), \dots, (s_t, a_t, b_t, \alpha_t)]$, where $s_i$ is the state, $a_i$ is the action taken by the policy or expert, $b_i$ indicates whether an expert intervention is triggered at timestep $i$, $\alpha_i$ is the expert's intervention threshold parameter at timestep $i$, and and $|\tau| = t+1$ is the length of the trajectory, with indices $i \in \{0, 1, \dots, t\}$.  For states with expert control, we assume optimal behavior and assume a cost of 0.

We sample trajectory segments $\sigma_i$ and $\sigma_j$ of length $w$ from the set of trajectories. Let $b(\sigma_i, \alpha_i) = 1$ if the final state of segment $\sigma_i$ triggers an intervention under threshold $\alpha_i$. Using these intervention indicators, we derive pairwise preference labels comparing the total true cost of segments $\sigma_i$ and $\sigma_j$.
\begin{equation}
y(\sigma_i > \sigma_j) =
\begin{cases}
\text{True}, & 
\text{if } 
\begin{aligned}[t]
& b(\sigma_i, \alpha_i) = 1, \\
& b(\sigma_j, \alpha_j) = 0, \\
& \alpha_i \ge \alpha_j,
\end{aligned} \\
\text{False}, & \text{otherwise.}
\end{cases}
\label{idea_1_ylabel}
\end{equation}

\begin{algorithm}[tb]
   \caption{Learning from Delayed Interventions(LfDI)}
   \label{alg:1}
\begin{algorithmic}
   \STATE {\bfseries Input:} Environment env
   \STATE {\bfseries Input:} Buffer of trajectories $B$, 
    \STATE {\bfseries Input:} Cost Model training step $L$
   \STATE {\bfseries Input:} Number of trajectories sampled $M$
   \STATE {\bfseries Input:} Maximum step per trajectory $N$
   \STATE {\bfseries Input:} $\pi$, $\pi^*$, $c^*(s, a)$
   \STATE {\bfseries Input:} Intervention window $w$
   \STATE {\bfseries Input:} Intervention Threshold $\alpha$
   \FOR{i in 1 ... M}
    \STATE: $a_1$ = $\emptyset$
   \FOR{t in 1 ... N}
        \STATE: $s_t \sim env(s_t, a_t)$
        \STATE: $a_t \sim \pi(s_t)$
        \IF{$\sum_{i=t-w}^{t} c^*(s_i, a_i) > \alpha$}
            \STATE: $a_{t+k} \sim \pi^*(s_{t+k}), \quad k = 0, \dots, x-1, \quad x < N$, $s_{t+k + 1}\sim env(s_{t+ k, a_{t + k}})$
        \ENDIF
        \STATE: $B \leftarrow B \cup \{(s_t, a_t, \alpha_t, b(\tau_t, \alpha_t))\}$   \ENDFOR
\STATE: $\pi \leftarrow \arg\min_{\pi}
- E_{(s,a)\sim B}
\big[ \log \pi(a \mid s) \big]$
   \ENDFOR
   \FOR{$k = 1$ \TO $L$} 
    \STATE $(\sigma_i, \sigma_j, y(\sigma_i > \sigma_j)) \sim B$    \STATE: $P(\sigma_i > \sigma_j) = \frac{ e^{\sum c(s_{i, t}, a_{i, t})}}{ e^{\sum c(s_{i, t}, a_{i, t})} + e^{\sum  c(s_{j, t}, a_{ij t})}}$
\STATE $c \leftarrow \arg\min_{c} \; \mathcal{L}_{\text{cost}} \; \text{(Eq.~\ref{cost_loss_idea_1})}$
   \ENDFOR
\STATE $\pi \gets \arg\min_\pi L_\pi \quad (\text{Eq.~\eqref{total_loss_idea_1}})$
\end{algorithmic}
\end{algorithm}
Put into words, when we sample a trajectory with intervention threshold $\alpha$, we classify the trajectory segments into segments that lead to expert intervention, $b(\sigma, \alpha)=1$, and segments without intervention $b(\sigma, \alpha)=0$.  Hence, a trajectory segment $b(\sigma, \alpha)=1$ will have a higher cost than any segments that do not lead to interventions, provided they have the same or lower intervention threshold $\alpha$. Trajectory segments with lower intervention thresholds and not leading to interventions are assumed to have lower cost, since a segment triggering intervention must have accumulated cost at least $\alpha$, while non-intervened segments with threshold $\alpha$ have less than $\alpha$ cost. Our preferences are not limited to comparisons within the same trajectory; our approach considers preferences between trajectory segments of different trajectories in the same manner. We propose training a cost function $c(s, a)$ that takes as input a state $s$ and action $a$ and outputs its estimated cost as a scalar. For each trajectory segment $\sigma_i$, we can obtain the probability of preferring one trajectory segment(e.g., with higher cost) over another in terms of cost with the Bradley-Terry model and train the model with the loss proposed by Christiano et al.\cite{christiano2017deep}.
\begin{equation}
\begin{aligned}
P(\sigma_i > \sigma_j) &= 
\frac{
    e^{\sum c(s_{i, t}, a_{i, t})}
}{
    e^{\sum c(s_{i, t}, a_{i, t})} + e^{\sum c(s_{j, t}, a_{j, t})}
}
\end{aligned}
\label{pref_prob}
\end{equation}

\begin{equation}
\begin{aligned}
L_{pref} = - (&\, y(\sigma_i > \sigma_j) \cdot \ln P(\sigma_i > \sigma_j) \\
& + y(\sigma_j > \sigma_i) \cdot \ln P(\sigma_j > \sigma_i) \,)
\end{aligned}
\label{pref_loss}
\end{equation}
Training the cost function with only our proposed preference loss (\ref{pref_loss}) can lead to poor performance, as we will later show in the evaluation, due to insufficient constraints on the cost function. Therefore, we propose to augment the loss function with a noisy expert cost loss, where we add noise to expert actions and train the cost function predict a cost equal to the magnitude of the additional noise. This loss can constrain the cost function, reducing the probability of learning unrealistic costs. This loss can only be applied to states with expert actions. 
\begin{equation}
L_{expert} = (c(s, a^* + \epsilon) - \|\epsilon\|)^2
\label{idea_1_noisy_expert_loss}
\end{equation}
where $a^* + \epsilon$ denotes an expert action with added noise $\epsilon$, and $\epsilon$ is randomly sampled and applied to the expert action. Our final loss for training the cost function is as follows:
\begin{equation}
    L_{cost} = \beta * L_{pref} +  L_{expert} + \lambda_{cost} \|\theta_{cost}\|_2^2
    \label{cost_loss_idea_1}
\end{equation}
Where $\beta$ is the hyperparameter of the weight of the preference loss. Here, $\theta_{cost}$ denotes the parameters of the learned cost function $c(s, a)$, and $\lambda_{cost}$  controls the strength of L2 regularization. 
\begin{equation}
    L_{\pi} = c(s, \pi(s))^2 + \|\pi(s) - a^*\|^2 + \lambda_{\pi} \|\theta_{\pi}\|_2^2
    \label{total_loss_idea_1}
\end{equation}
The overall loss for training the policy $\pi$ after training the cost function consists of minimizing the cost of the policy's predicted actions on all states, the behavior cloning error on expert states, and an L2 regularization.  Here, $\theta_{\pi}$ denotes the parameters of the policy $\pi$, and $\lambda_{\pi}$ controls the cost regularization trade-off between avoiding high-risk pre-intervention states and preserving task progress (overly high costs encourage conservatism). 

Our overall algorithm is described in Algorithm~\ref{alg:1} and Figure~\ref{delayed_intervention_data_sampling}. We assume a pretrained but suboptimal initial policy $\pi$. For each trajectory, the expert is provided a target intervention threshold $\alpha$. The expert intervenes when the accumulated cost within $w$ exceeds $\alpha$. After an intervention is triggered, the expert takes control until it is safe to return control to the policy. After each trajectory, we retrain the sampling policy on new intervention data. The target intervention threshold increases over time as the policy improves. Once we complete intervention data collection, we train the cost function $c$ with Equation~\ref{cost_loss_idea_1}. For the $L_{\text{pref}}$ term in Equation~\ref{pref_loss}, we sample one segment without intervention and one segment with intervention. Once we complete training the cost function, we train the policy $\pi$ using behavior cloning on states with expert actions and the learned cost function on all states using Equation~\ref{total_loss_idea_1}.

\section{Evaluation}
\label{sec:evaluation}
\begin{figure}
\includegraphics[scale=0.6]{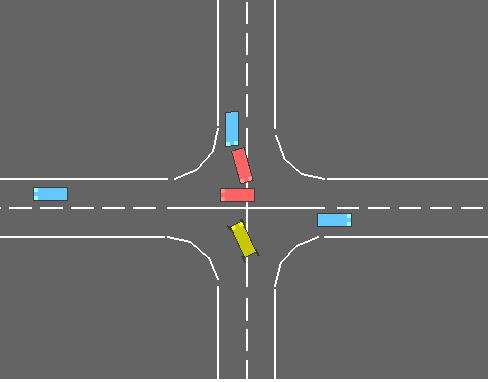}
    \caption{ We evaluated our method on the \textbf{Intersection} environment from highway-env~\cite{highway-env}. In this environment, the agent must drive north to the intersection and make an unprotected left turn. Due to the limited vehicle logic in highway-env, other vehicles can get stuck and crash into each other. Consequently, the overall success rate for even an expert is not perfect, making it critical for the agent to learn when to stop and when to turn safely.} 
    \label{fig:intersection}
\end{figure}
We apply our HG-Dagger-based imitation learning framework described in Algorithm~(\ref{alg:1}) to sample trajectories with expert interventions. Initially, we assume that the expert begins with a low intervention threshold $\alpha$, and therefore low intervention delay. This reflects the likelihood that the sampling policy performs poorly initially and the expert intervenes frequently. After sampling a trajectory, we retrain the policy and continue collecting additional expert interventions. Once a fixed number of expert interventions has been sampled, we increase the target intervention threshold $\alpha$. This simulates the idea that, as the policy improves, the expert begins to trust it to recover from minor errors and becomes less attentive.

For environments with human experts, we query the expert after each intervention for a more precise $\alpha$; the expert provides their actual intervention threshold, treating the target threshold as guidance instead. We use the target threshold as a noisy label in our ablation evaluations in table~\ref{tbl:highway_ablation_results}. During data sampling, we found that intervention delays occurred naturally while the expert supervised the agent, even when we do not consider the target threshold. There were multiple occasions when the expert simply did not react in time. In other cases, the expert could see the agent slowly making a mistake but might have waited to see if the agent can recover on its own. In some cases, the expert took control, but it was too late to prevent the agent from crashing. We note that the expert does not need to be queried immediately after intervention. We can query the expert after all data is collected, similar to human preference learning. When querying human experts for $\alpha$, we use the following guidelines:
\begin{itemize}
    \item $\alpha = 1$: The agent makes a minor mistake, such as a slight slowdown or a small deviation from optimal behavior.
    
    \item $\alpha = 2$: The agent makes a significant mistake that will lead to a crash within 10 steps if no intervention occurs, slows down substantially, or misses a clear opportunity to complete the task.
    
    \item $\alpha = 3$: The agent is in a critical state and will terminate unless immediate intervention is applied.
    
    \item $\alpha = 4$: The agent has already crashed, or the expert has taken control but is unable to prevent termination.
\end{itemize}
\begin{figure}
\includegraphics[width=\columnwidth,keepaspectratio]{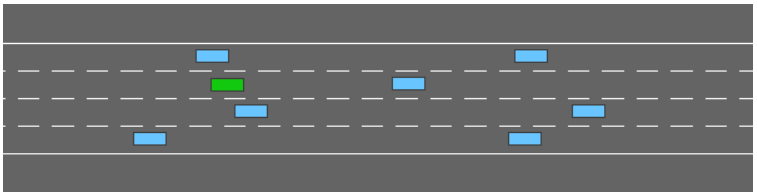}
    \caption{ The \textbf{Highway} environment from highway-env~\cite{highway-env} consists of a long 4-lane highway with a controlled vehicle. The objective is to travel as far (in terms of x-position) as possible within a time limit of 500 steps without crashing or leaving the road.}
    \label{fig:highway}
\end{figure}
To reduce cognitive load, LfDI uses four coarse delay thresholds rather than exact numerical values, making annotations lightweight while preserving the ordinal structure needed for preference learning. Finer labels might improve performance but would increase expert burden and annotation error; future work could study this trade-off in physical settings and automate delay estimation. Finally, we also sample additional trajectories without expert supervision; terminations are automatically assigned an intervention threshold of $\alpha=4$. We compare our method with and without these additional data to show that our approach can leverage unsupervised trajectories without expert intervention and still improve performance, as shown in Table~\ref{tbl:highway_ablation_results}. When training our method and all others methods, we use both purely expert-controlled trajectories and expert intervention data from our human-gated imitation learning system. For all other methods that train a cost function, we can also use the cost function to minimize on sampled states without expert data. 

\subsection{Baselines and Comparisons}
We compare against HG-DAgger, a noisy expert cost function, and EIL~\cite{spencer2020learning}. HG-DAgger uses the expert intervention data naturally collected by our human-gated system. The noisy expert is trained with loss~(\ref{idea_1_noisy_expert_loss}) on expert-only data, and the policy with loss~(\ref{cost_loss_idea_1}) on all data. For EIL, we tested windows of 5 and 30 timesteps before interventions to identify bad states. Labeling all other states as good, or labeling all states within the 30-step window as bad, both performed poorly---suggesting the window contains mixed-quality states that cannot be equally penalized. We therefore label only states outside the 30-step window as good.

For the human preference learning baseline, we use a combination of a noisy expert and preference learning to train the cost function. Experiments using only preference learning performed poorly; this may be due to an insufficient number of trajectory segment preference pairs. For this baseline, we assumed access to a limited number of fixed preference queries. To obtain these preferences, we first trained an ensemble of 5 behavior cloning policies, each trained on a randomly sampled 80\% subset of the data buffer, including both initial expert data and intervention data. We then computed an ensemble uncertainty for each trajectory segment without expert data using the standard deviation of the predicted actions. Segments were sorted by mean ensemble uncertainty, and we randomly selected pairs from the top-k highest-uncertainty segments for querying. This strategy follows prior work by Lee et al.~\cite{lee2021pebble}, and Christiano et al.~\cite{christiano2017deep}. Preferences were obtained either from the oracle function~(\ref{oracle_cost}) or queried directly from human experts. In terms of the number of preference queries, we used at least as many preference queries as expert interventions for fair comparison. For all our tables and figures, the $\pm$ is 1 standard deviation.

\begin{figure}
\includegraphics[scale=0.9]{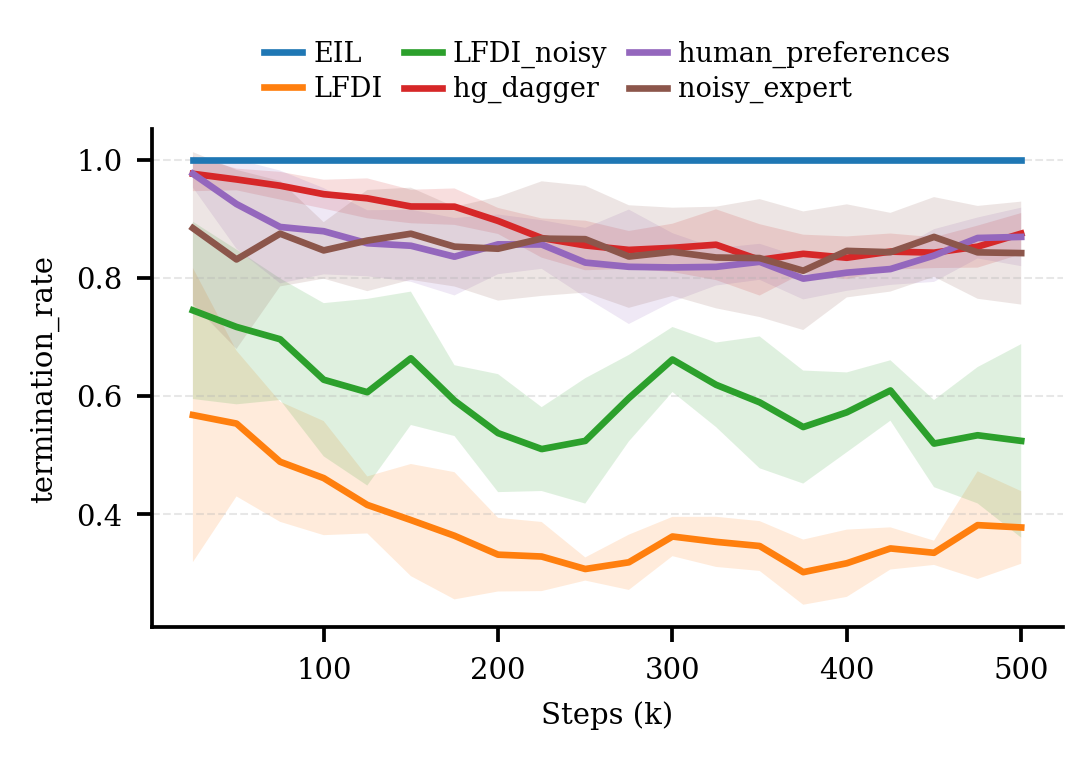}
    \caption{We plot the average Highway environment termination rate for our method(LFDI) with and without noisy labels, EIL, noisy expert, and HG-Dagger. LFDI\_noisy is a policy trained with LFDI and noisy and imprecise supervision-level labels.} 
    \label{fig:highway_plot}. 
\end{figure}
\begin{table}[h]
\centering
\setlength{\tabcolsep}{3pt}
\label{tbl:highway_env_results}
\begin{tabular}{@{}lccc@{}}
\toprule
\textbf{Method} & \textbf{Dist / Task Comp.} $\uparrow$ & \textbf{Term.} $\downarrow$ & \textbf{Speed} \\
\midrule
\multicolumn{4}{l}{\textbf{\textit{Highway}}} \\
\addlinespace[2mm]
EIL & $129.7 \pm 9.0$ & $1.00 \pm 0.00$ & $\mathbf{2.77 \pm 0.11}$ \\
LfDI & $\mathbf{927.7 \pm 55.8}$ & $\mathbf{0.34 \pm 0.06}$ & $2.23 \pm 0.09$ \\
HG-DAgger & $509.5 \pm 47.7$ & $0.85 \pm 0.04$ & $2.34 \pm 0.10$ \\
human-pref & $512.7 \pm 32.1$ & $0.83 \pm 0.05$ & $2.30 \pm 0.16$ \\
noisy-expert & $381.4 \pm 70.0$ & $0.84 \pm 0.08$ & $2.18 \pm 0.32$ \\
\midrule
\multicolumn{4}{l}{\textbf{\textit{Roundabout}}} \\
\addlinespace[2mm]
EIL & $0.610 \pm 0.074$ & $0.299 \pm 0.069$ & -- \\
HG-DAgger & $0.612 \pm 0.050$ & $0.362 \pm 0.050$ & -- \\
LfDI & $\mathbf{0.722 \pm 0.041}$ & $\mathbf{0.236 \pm 0.042}$ & -- \\
human-pref & $0.654 \pm 0.0491$ & $0.319 \pm 0.059$ & -- \\
noisy-expert & $0.563 \pm 0.050$ & $0.401 \pm 0.065$ & -- \\
\midrule
\multicolumn{4}{l}{\textbf{\textit{Intersection}}} \\
\addlinespace[2mm]
EIL & $0.481 \pm 0.053$ & $0.438 \pm 0.039$ & -- \\
HG-DAgger & $\mathbf{0.520 \pm 0.035}$ & $0.449 \pm 0.035$ & -- \\
LfDI & $0.513 \pm 0.048$ & $\mathbf{0.399 \pm 0.039}$ & -- \\
human-pref & $0.505 \pm 0.032$ & $0.455 \pm 0.030$ & -- \\
noisy-expert & $0.500 \pm 0.032$ & $0.453 \pm 0.032$ & -- \\
\bottomrule
\end{tabular}
\caption{Results for three highway-env environments. Highway shows distance traveled[sim. units] where the distance is in simulated units, termination rate(lower is better), and speed[sim. units/s]. Roundabout and Intersection show task completion rate instead and termination rate.}
\end{table}
\subsection{Highway-env Results}
For our evaluation with human experts, we used the highway-env~\cite{highway-env}, a 2D environment simulating a multi-lane road system with traffic. The agent is controlled by a continuous action \(a = [\text{acceleration}, \text{steering}]\). This simple action space makes the environment suitable for human experts control. To increase difficulty, we raised the maximum acceleration and speed limit and increased the number of vehicles on the road.

We evaluated the agent in three environments: Highway (Figure~\ref{fig:highway}), Intersection (Figure~\ref{fig:intersection}), and Roundabout (Figure~\ref{fig:roundabout}). The Highway environment is a 4-lane road with multiple vehicles moving at varying speeds; performance is measured by distance traveled within a time limit. The Roundabout environment requires entering the roundabout and exiting at the second exit, while the Intersection environment requires completing an unprotected left turn. For both Roundabout and Intersection, performance is measured by task completion (e.g., reaching a designated road edge).

We also report the termination rate, triggered when the agent leaves the road, drives on the wrong side of the road, or collides with other vehicles, as a measure of safety. For human preference learning, we collected \(2{,}200\), \(800\), and \(1{,}000\) trajectory pairs for Highway, Roundabout, and Intersection, respectively, matching the number of interventions in our method to ensure fair comparison. Note that in the Roundabout and Intersection environments, it is possible for an agent to neither terminate nor complete the task.
For the Highway environment, our method achieves significantly better performance than all other methods, as shown in Table~\ref{tbl:highway_env_results}, Table ~\ref{tbl:highway_ablation_results}, and Figure~\ref{fig:highway_plot}, while the remaining methods perform comparably. We also observe that using noisy intervention threshold labels degrades performance, highlighting the importance of high-quality queried labels. Nevertheless, even with extremely noisy labels, our approach outperforms all other methods.

Our method explicitly accounts for delayed interventions and is therefore more effective at avoiding risky behaviors, leading to substantially lower termination rates and allowing the agent to travel farther compared to alternative approaches. As a tradeoff, our method exhibits slightly lower average speed. In contrast, EIL demonstrates the opposite behavior: it achieves the highest average speed but consistently terminates. EIL promotes states that do not lead to interventions, which would work if expert interventions were not delayed. However, with delayed interventions, this assumption introduces errors, resulting in an overly optimistic cost function that could lead to unsafe behavior. Another factor for our method's performance over other methods is that due to poor initial HG-DAgger performance, we sampled over 2200 expert interventions, the most of all our experiments. Our method learns from interventions and therefore is expected to do better with more delayed intervention signals.  
\begin{table}[h]
\centering
\label{tbl:highway_ablation_results}
\begin{tabular}{lccc}
\hline
\textbf{Method} & \textbf{Distance} $\uparrow$ & \textbf{Term.} $\downarrow$ & \textbf{Speed} \\
\hline
LfDI & $928.2 \pm 55.6$ & $0.34 \pm 0.06$ & $2.23 \pm 0.09$ \\
LfDI(noisy labels) & $705.7 \pm 104.5$ & $0.58 \pm 0.10$ & $2.20 \pm 0.12$ \\
LfDI(no unsup) & $832.9 \pm 36.9$ & $0.50 \pm 0.07$ & $2.29 \pm 0.08$ \\
LfDI(no noisy) & $380.9 \pm 30.2$ & $0.99 \pm 0.00$ & $3.01 \pm 0.06$ \\
LfDI(ensemble) & \textbf{$970.0 \pm 24.8$} & \textbf{$0.28 \pm 0.04$} & $2.22 \pm 0.06$ \\
\hline
\end{tabular}
\caption{Highway Ablation Study Results (Last 10 Steps). We provide a set of results of our method with different ablation settings. Note that distance and speed are in simulator units. }
\end{table}

Furthermore, we evaluated our method with additional ablation tests. First, we tested our method with noisy intervention labels, where instead of using the queried intervention label given by the expert, we used the guidance intervention threshold as the label. There is a significant performance gap between our method with clean labels and noisy labels, but our method still outperforms all other methods for Highway. Next, we trained our policy without additional unsupervised sampled trajectory data. From Table~\ref{tbl:highway_ablation_results}, we can see that the additional unsupervised data can help improve performance, but our method still significantly outperforms all other methods. 

Additionally, we experimented with training our method with an ensemble of cost functions(5) trained without the noisy expert loss~\ref{idea_1_noisy_expert_loss}; the results are quite bad and show the need for the noisy expert loss. Finally, we experimented with training our policy with an ensemble of cost functions (5) instead of 1 cost function. We can see from the results that using an ensemble slightly improved performance and reduced the standard deviation. This is expected as training with an ensemble of cost functions is more stable and could prove useful for environments with less expert intervention data. Training with an ensemble of cost functions can be done for any method with a cost function(EIL, noisy expert, human preference), and we expect similar performance improvements for other methods when trained with a cost ensemble.

In the Roundabout environment, our method achieves the best performance in both task completion rate and termination rate, as shown in Table~\ref{tbl:highway_env_results}. Human preference learning ranks second, followed by EIL and HG-Dagger, while the noisy expert baseline performs the worst.

\begin{figure}
\includegraphics[scale=0.9]{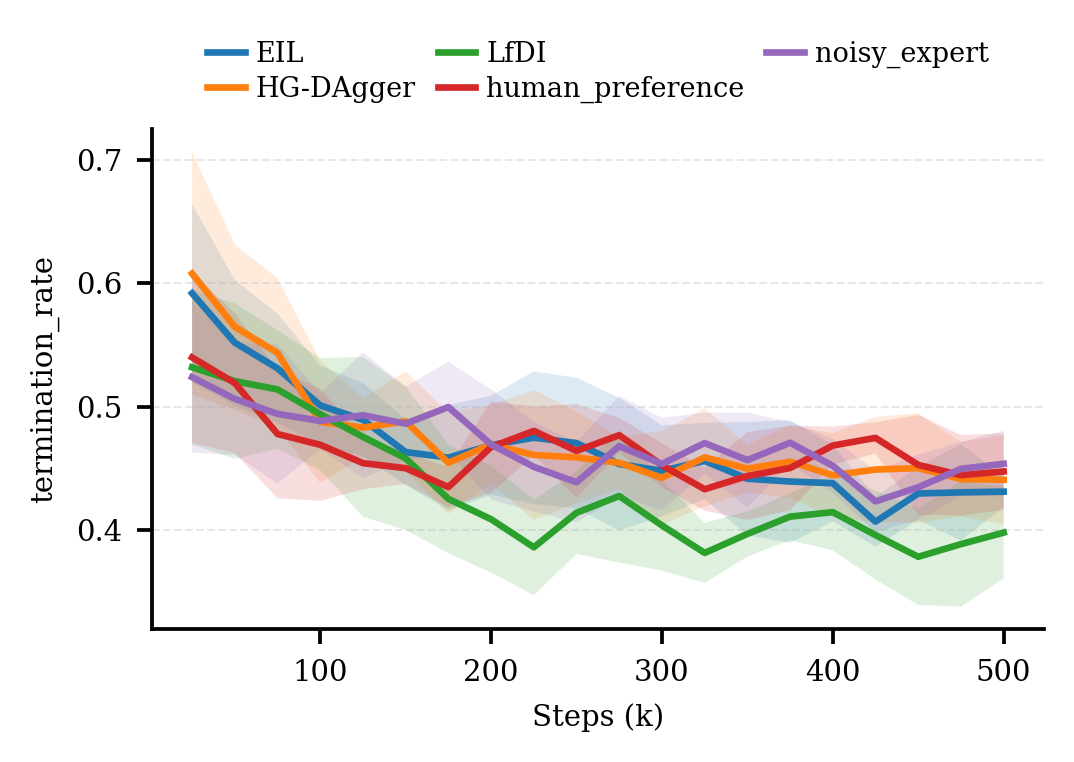}
    \caption{We plot the termination rate (crash rate) of our method(LFDI) against other methods. We can see that our method shows improvements in reducing crash rates without decreasing completion rates, as shown in Table~\ref{tbl:highway_env_results}. } 
    \label{fig:intersection_result_termination}. 
\end{figure}

In the Intersection environment, task completion rates are relatively similar across all methods, with HG-Dagger, our method, and human preference learning achieving comparable performance. One possible explanation is that the task completion rate is effectively capped by the environment dynamics: the limited behavioral logic of non-controlled vehicles often leads to crashes that block the road, as illustrated in Figure~\ref{fig:intersection} preventing task completion. This means that in cases where the road is blocked, the agent must avoid crashing into the blocked vehicles. Our method achieves lower termination rates than all other baselines while maintaining the same task completion rate as other methods(Figure~\ref{fig:intersection_result_termination}), highlighting the benefit of learning from delayed interventions.

\begin{table}[h]
\centering
\label{tbl:humanoid_results}
\begin{tabular}{lcc}
\toprule
\textbf{Method} & \textbf{Reward} $\uparrow$ & \textbf{Term.} $\downarrow$ \\
\midrule
\multicolumn{3}{l}{\textit{\textbf{High Delay Dataset}}} \\
\addlinespace[2mm]
EIL & $2433 \pm 282$ & $15.10 \pm 3.18$ \\
HG-DAgger & $3344 \pm 324$ & $6.68 \pm 1.89$ \\
LfDI & $\mathbf{3863 \pm 310}$ & $\mathbf{3.99 \pm 1.34}$ \\
human-pref & $3739 \pm 324$ & $4.36 \pm 1.54$ \\
noisy-exp & $3587 \pm 353$ & $5.56 \pm 1.92$ \\
\midrule
\multicolumn{3}{l}{\textit{\textbf{HG-DAgger Intervention Delay Experiment}}} \\
\addlinespace[2mm]
HG-DAgger high delay & $3343 \pm 323$ & $6.69 \pm 1.89$ \\
HG-DAgger avg delay & $3714 \pm 310$ & $4.33 \pm 1.79$ \\
HG-DAgger low delay & $\mathbf{4165 \pm 194}$ & $\mathbf{2.14 \pm 0.72}$ \\
\bottomrule
\end{tabular}
\caption{Humanoid Environment Results (Last 10 Steps). Humanoid results (top) show all methods with limited expert interventions. For the bottom, we performed a simple experiment showing the performance of HG-DAgger with different levels of expert intervention delays.}
\end{table}

\subsection{MuJoCo Results}
We also evaluate on MuJoCo Humanoid, whose action space is far more complex than highway-env’s two-dimensional one. Because Humanoid is too difficult for human experts to control and intervene in, we follow common practice in this domain \cite{lee2021pebble} and use an oracle cost function with an expert policy to determine when interventions occur.  Our oracle cost function is defined as follows, where $\pi^*(s)$ is the expert policy:
\begin{equation}
    c^*(s, a) = (\pi(s) - \pi^*(s))^2
    \label{oracle_cost}
\end{equation}
Table~\ref{tbl:humanoid_results} shows that LfDI slightly outperforms the baselines when trained with only 150 expert interventions. The performance gap is limited because, with more interventions, all methods can easily approach expert performance in Humanoid, leaving little room for meaningful comparison. Thus, these results show that LfDI remains useful under limited expert data. We further evaluate HG-DAgger under different intervention delays using equal-size expert datasets, where the dataset label reflects the average delay. Humanoid is used for this study because the oracle cost function provides more precise intervention-threshold queries than human experts. The results show that high-delay data performs the worst, average-delay data performs next, and low-delay data performs the best, highlighting the impact of intervention delay on policy performance and the need to account for delayed intervention data properly.

In conclusion, this work studies an often-overlooked source of information in interactive imitation learning: delayed human interventions. Through experiments and ablations, LfDI shows that intervention delays can provide a useful learning signal and improve policy learning under limited intervention data. These results suggest that intervention delay should be explicitly considered in future IIL methods.

\addtolength{\textheight}{-12cm}   




%
%
\bibliographystyle{IEEEtran}
\bibliography{sample}

\end{document}